# The Grasps Under Varied Object Orientation Dataset: Relation Between Grasps and Object Orientation*

Chang Cheng, Yadong Yan, Qing Yang, Mingjun Guan, Jianan Zhang, and Yu Wang, *Member, IEEE*

*Abstract*—After a grasp has been planned, if the object orientation changes, the initial grasp may not have to be modified to accommodate the orientation change. For example, rotation of a cylinder by any amount around its centerline does not change its geometric shape relative to the grasper. Objects that can be approximated to solids of revolution or contain other geometric symmetries are prevalent in everyday life, and this information can be employed to improve the efficiency of existing grasp planning models. This paper experimentally investigates change in human-planned grasps under varied object orientations. With 13,440 recorded human grasps, our results indicate that during pick-and-place task of ordinary objects, stable grasps can be achieved with a small subset of grasp types, and the wrist-related parameters follow normal distribution.

## I. INTRODUCTION

Grasping is one of the most fundamental yet sophisticated tasks of the human hands. For simply picking up an object, a grasp requires object shape recognition, initial grasp planning, and sensory-motor control feedback from the human [1]. Despite many existing robotic hands can nearly restore the Degree of Freedom (DOF) of human hands, the lack of adequate grasp policies and sensory feedback hampers the attempt to plan and produce stable grasps.

Current grasp syntheses are classified into analytical and empirical ones [2]. The analytical approaches, such as [3] [4] [5], examine a mainly mathematical and computational problem [2]. Empirical grasp synthesis, such as [6], [7], [8], [9], are experience-based, and typically combines human grasp data with robotic learning policies to transfer observed grasps [10]. For both approaches, the complexity of anthropomorphic hands complicates grasp planning. Recent works have frequently modeled a human hand with over 20 DOF [11] [12]; with 6 additional DOF to define hand position and orientation, the grasp configuration space contains nearly 30 dimensions when fully parameterized. This high dimensionality calls for simplification.

In the discrete sense, the seminal work by Napier [13] divided common grasps into "precision" and "power" grasps, which is expanded into 16 more specific types by Cutkosky [14], and later into 33 by Feix et al. [15]. These classifications simplify the process of transferring human grasps to robots for grasp synthesis. For example, Ekvall et al. constructed a grasp planning system where a grasp pre-shape is provided, and the robot searches for an approach vector to complete the grasp [16]. Using algebraic methods, Santello et al. [17] used data-gloves to measure joint angles during grasps and represented postural synergies in human handsc as eigenvectors. This reduced the hand configuration space to 3 dimensions while preserving 70% of individual joint variation. These synergies have shown success in reducing the control complexity of both fully actuated [18] and underactuated [19] anthropomorphic hands. Building on the work of Santello et al., [20] and [21] have developed grasp planning algorithms in reduced postural spaces. While these two algorithms are based on different frameworks: simulated annealing and reinforcement learning, both algorithms demonstrated that stable grasps can be planned and optimized in significantly reduced grasp subspaces.

Aside from finger-specific parameters, wrist-related parameters are also essential to a grasp. Balasubramanian et al. showed that when human-planned grasps outperform computer-planned ones, the difference often lie in the wrist orientation [22]. When postural synergies are used to parameterize hand postures, a grasp is often described by 9 variables: 3 for eigenvalues and 6 for wrist position and orientation. This implies that searching efficiently in the wrist transformation space is as important as in joint parameter space in building efficient grasp planning algorithms. Perhaps the most practical task for grasp algorithms is to pick up and place an object from a planar surface. To achieve such task, most existing algorithms would determine a maximum distance the hand can be away from the object, then search for a wrist transformation within that distance. One property of these algorithms is they do not consider rotational symmetry within the objects, which can be used to reduce the wrist transformation search space. Consider when a human reach for a cylinder. If the cylinder rotates about its vertical axis, the initial grasp does not need to change to maintain a stable grasp. This phenomenon is illustrated in Fig. 1, where we asked a volunteer to grasp a tennis ball, a stapler, and a screwdriver under different object orientations. When the objects rotate around the axis orthogonal to the plate by 45 degrees, the human hand makes little adjustments to reproduce a stable grasp. By including every angle around the object in the search space, the traditional search parameters may be addressing a more difficult task than needed. In this paper, we aim to

*Research supported by "National Key R&D Program of China" under grant 2017YFA0701101.

Y. Yan, M. Guan, J. Zhang, and Y. Wang (corresponding author) are with the School of Biological Science and Medical Engineering, Beihang University, Beijing 100191, China (fax: 8610-82315554; email: adam7217@qq.com; yangq@buaa.edu.cn; mingjunguan7@163.com; baby0303zjn@buaa.edu.cn; wangyu@buaa.edu ).

C. Cheng is with the School of Biological Science and Medical Engineering, Beihang University, Beijing, China and the Department of Mathematics and Computer Science, Colorado College, CO 80946, USA (email: d_cheng@coloradocollege.edu)

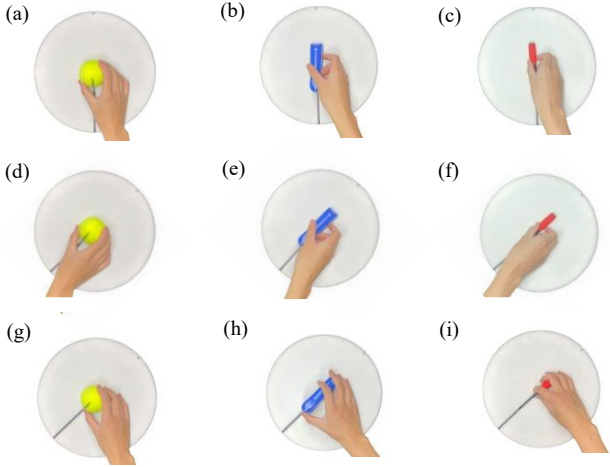

Fig. 1. Human grasps of objects under different orientations. A tennis ball (a), (d),(g), a stapler (b), (e), (h), and a screw driver (c), (f), (i) are grasped respectively. Grasps in (a-c) represent the grasps for the object in initial orientation, (d-f) represents the grasps where the hand rotates with the object, (g-i) are grasps that humans tend to choose for the new object orientation.

quantify the rotational variances needed in approach vectors to achieve stable grasps. Additionally, we seek to provide a reference for the radius of the grasp search space. Due to the complexity that arises from searching, most grasp planners employ heuristics to create a bounding box or sphere for the planner. For example, in the GraspIt! simulation environment by Miller, Allen, and Ciocarlie, the hand is bounded within a box with width of 500mm [23, 25]. The system developed by Bohg et al. aligned the approach vector of the hand with the object surface normal and limited the approach distance in a fixed range of 0-200mm [24]. While studies like these have yielded meaningful results toward the field of grasp analysis, the specific parameter values they employed are not empirically validated. To the authors knowledge, there does not exist a study that rigorously quantifies these parameters. In grasp planning systems, parameters related to wrist transformation, such as how far the hand can be away from the object, can make or break the system. In the past, these parameters have been based on researchers' logical inference. Providing empirical reference for choosing such parameters in future studies is another one of this paper's goals.

In this paper, we experimentally explore change in human grasps for everyday objects with respect to change in object orientation. Such relation is organized into the Grasps Under Varied Object Orientation Dataset (GUVOOD). We then present our results to show that a small subspace of the wrist transformation and hand configuration is enough for grasping most ordinary objects.

## II. MATERIAL AND METHODS

This section describes the methods employed to investigate the relation between human-planned stable grasps and object orientation. Human subjects are instructed to grasp everyday objects under different orientations, and the change of the grasps with respect to changes in object orientation is measured. The experimental setup is shown in Fig. 2. An optical tracking system is employed to record the spatial transformation of the wrist, and two cameras are employed to register the hand posture. Therefore, any observed grasp is parameterized by seven variables: $\varphi$, $\theta$, $r$, $Rx$, $Ry$, $Rz$, $t$; whereas $\varphi$, $\theta$, and $r$ describe the wrist position in polar coordinates, $Rx$, $Ry$, and $Rz$ describe the wrist rotation, and $t$ is the manually labeled grasp type from the Cutkosky grasp classification [14]. For each subject, the objects are rotated to and grasped under 8 different orientations that are evenly spaced.

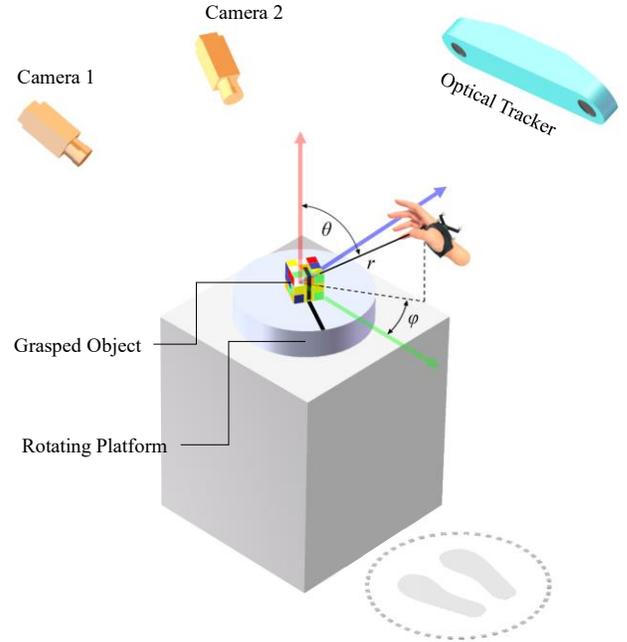

Fig. 2. Instruments in the experimental setup. The grasper stands opposite to the two cameras during the experiment.

### A. Grasped Objects

The object selection aims to include daily objects with varying sizes and shape complexities. We surveyed a panel of five individuals for thirty objects they frequently interact with. After excluding repeats and objects too large for our experiment, a total of 60 objects are chosen and are displayed in Fig. 3. Among our object set, 21 and 15 have counterparts from [17] and the YCB dataset. It is worth noting that some objects have alternative ways of being placed on a surface; Fig. 3. displays the unique way each object is presented to the graspers.

Certain objects with symmetrical properties would be isomorphic to itself under other orientations. Take a tennis ball, for example; all rotations of a tennis ball yield the same geometric properties relative to the grasper. To investigate the effect of geometric symmetry on human-produced grasps, we categorize the grasped objects into three groups: solids of revolution (SoRs), cuboids, and irregular objects. SoRs refer to objects whose 3-D model can be obtained through rotation of a plane curve around a centerline orthogonal to the table, i.e., spheres and cylinders. The cuboids are objects that can be approximated to cubes or rectangular boxes. The irregular objects are neither SoRs nor cuboids. Some objects, such as a marker, would be SoRs if placed in alternative fashion than presented in Fig. 3. Because the rotating platform rotates the objects around the vertical centerline, objects without a vertical centerline as presented in Fig. 3 would not be considered as SoRs.

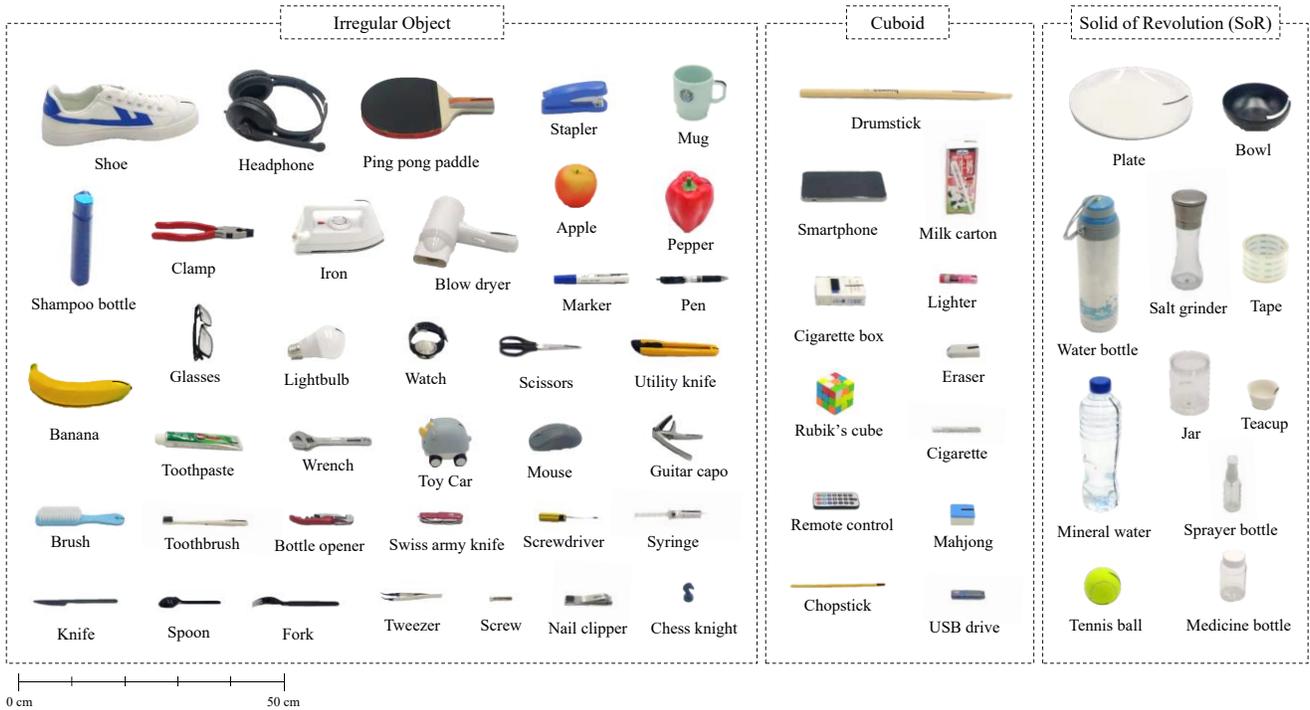

Fig 3. Images of actual objects used for the experiment. All objects are drawn to scale. The objects are classified into three categories by degree of symmetry: irregular objects, cublids, and solids of revolution (SoRs). For many of the objects above, there are multiple ways to place them on a surface. For example, a bowl can be placed with the opening facing up or down. In this experiment, the ways the objects are presented above are how they are placed during the experiment.

### B. Protocol

Previous work in the field has shown that grasps are strongly correlated with the usage of the objects [13]. This phenomenon may become a disturbance to this study. For instance, a syringe is commonly held with the index and middle finger on the barrel and the thumb at the end of the plunger for injection, but this is not the most stable and convenient way to grasp a syringe. To avoid potential bias occurring from object application, we instruct the subjects to neglect the conventional usage of an object and concentrate explicitly on the task of pick-and-place before experiments begin.

The subject sits roughly 0.4 meters away from the center of the rotating platform. The experiment begins as an operator puts the first object in the center of the platform. The subject is instructed to grasp the object and attempts a pick-up to demonstrate the success/failure of the grasp. The operator records the grasp along with the evaluation, and the subject places the object back on the platform. The operator rotates the platform by 45 degrees counterclockwise and repeats the steps above under the updated object orientation. Once grasps under all eight rotational orientations have been recorded, the next object is placed on the platform. This process repeats until grasps of all 60 objects have been recorded. Figure 4. shows sample data collected from one subject.

### III. RESULTS

Throughout the experiments, twenty-one righthanded and seven lefthanded subjects participated, and 13440 grasps were added to the GUVOOD overall. Among all grasps, 203(1.51%) contain ambiguous wrist transformation, and 86(0.63%) grasps failed. The ambiguous grasps resulted from the loss of sight of the reflective marker, and the failed grasps mainly occurred on flat and heavy objects such as wrench (48 fails) and knife (20 fails). This section presents and discusses the unambiguous, successful grasps in the GUVOOD.

### A. Wrist Position

A scatter plot of the palm positions in the GUVOOD is shown in Fig. 5 (a, b). For both left and right-handed graspers, the grasps congregate in a single cartesian octant. The center of the point clouds also exhibits a higher density of grasps. Fig. 5(e). displays the distribution of the $\varphi$, $\theta$, and $r$ used to parameterize grasp locations. The $\theta$, and $r$ distributions of the left and right-handed graspers are near identical and the $\varphi$ variables resemble the reciprocal of the other. The frequencies of all three position parameters bear resemblance to the standard normal probability density function (PDF) calculated from the mean and standard deviation (STD) of the variables except the $r$ frequency is slightly left-skewed. A possible cause of the skewness is that the reflective tracker is put on the back of the hand. Because $r$ essentially measures the distance from the hand to an object, and all grasps are made where the palm faces the object, the $r$ parameter becomes slightly biased by the thickness of the palm.

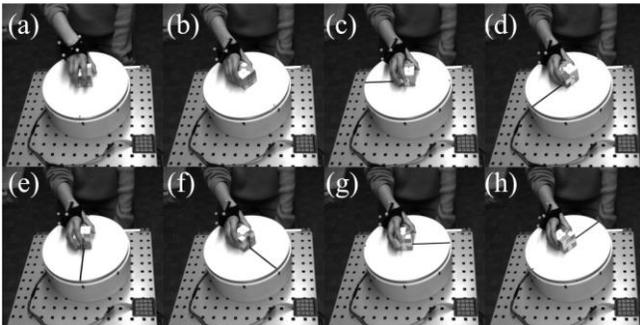

Fig 4. Example images of the recorded grasps under 8 different object orientations during the experiment.

## B. Wrist Orientation

Because the left and right-hand mirror each other, two different coordinate systems are built for the left and right hand to allow more comparable data representation, shown in Fig. 5(c, d). Both coordinate systems have the x-axis pointing above the back of the hand. For the left hand, the z-axis points towards the thumb and the y-axis points towards the wrist. For the right hand, the y-axis points to the thumb, and the z-axis to the wrist. The $Rx$, $Ry$, $Rz$ rotations are then calculated in XZY order for the left hand and XYZ order for the right. Similar to the frequency of the positional parameters, the wrist orientation parameter frequencies for the left and right-hand mirror each other, and the distribution resembles standard normal PDF.

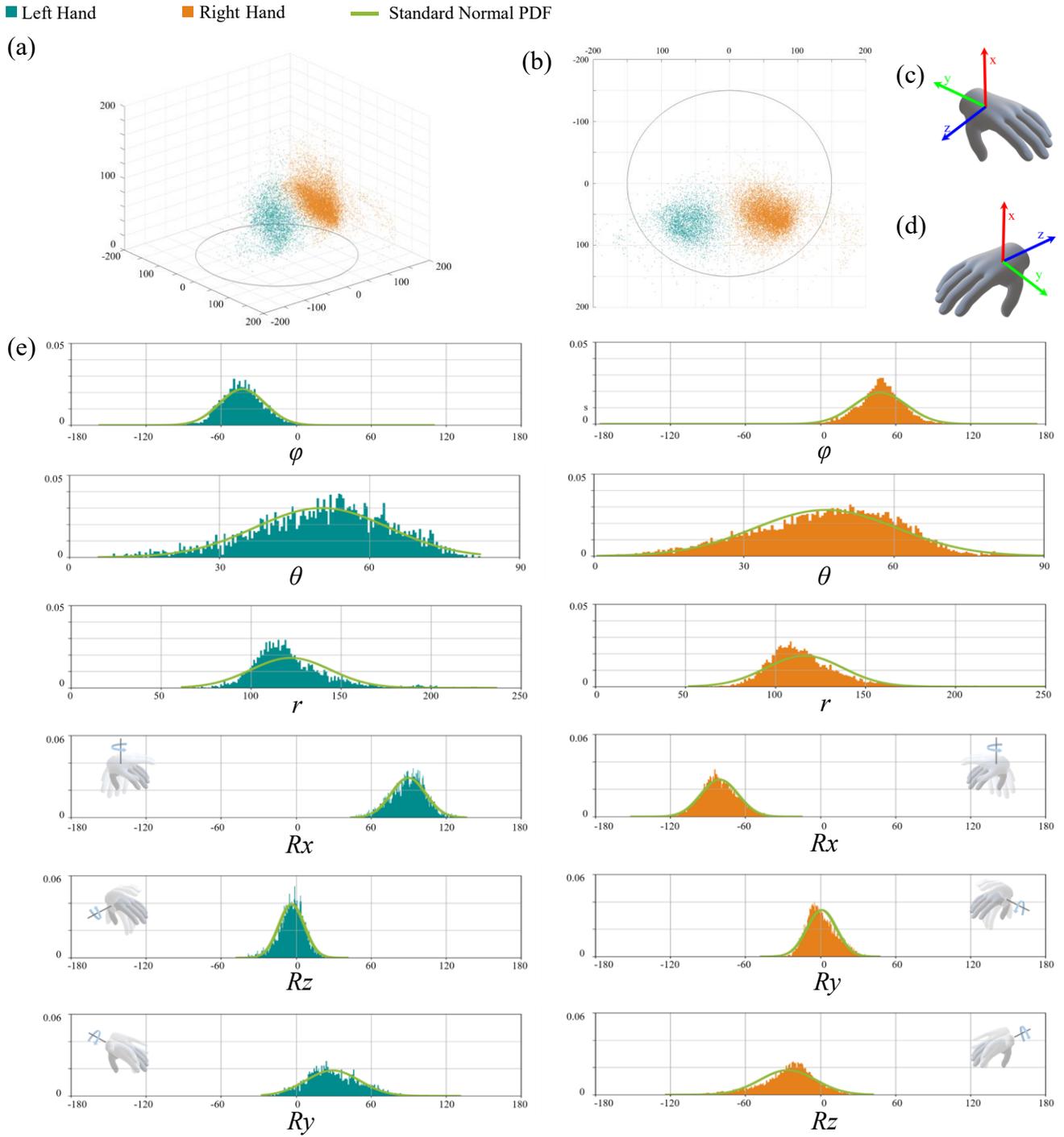

Fig 5. Distribution of the wrist transformation parameters from the grasp dataset, the graph legends are displayed above (a). (a) and (b) are scatterplots of the wrist positions, (c) and (d) display the coordinate system used for the left and right hand, respectively. (e) shows the frequency graphs for the corresponding parameters.

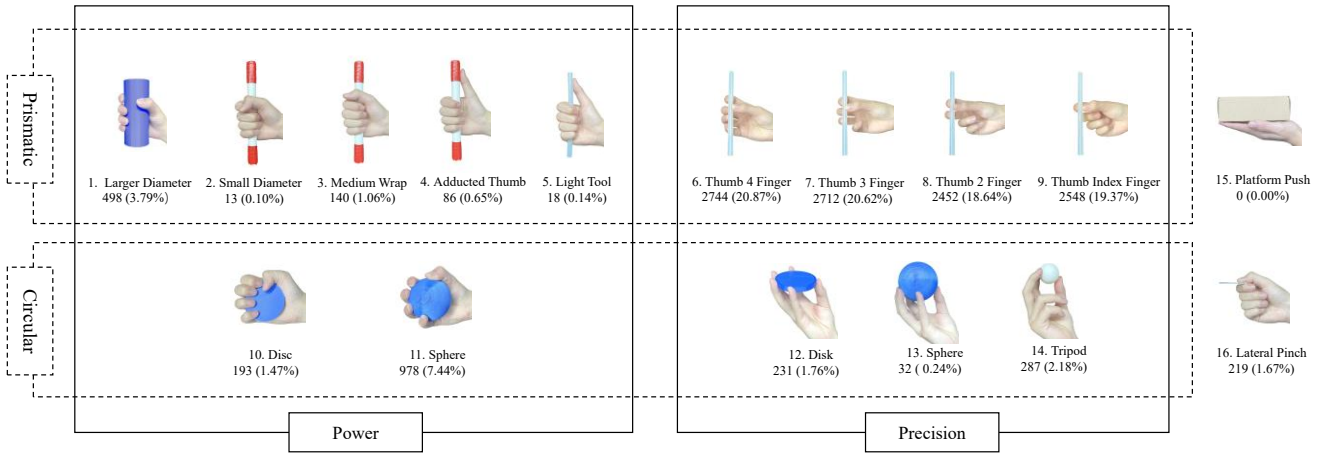

Fig 6. Distribution of grasp types observed in the dataset. The grasps that failed or could not be registered have been excluded. The number below each grasp type indicates the frequency, and the number in parentheses is the frequency as a percentage of total observations.

### C. Grasp Types

The grasps produced in the experiment are manually classified by the taxonomy proposed in [14], and the results are shown in Fig. 6. The most frequent grasps observed are the four precision prismatic grasps, which together account for 79.5% of total grasps. Precision and power grasps each make up 85.4% and 14.6% of the dataset. Such dominance of precision over the power grasps is more drastic than a comparable study done by Cini et al., which collected 62.0%, 7.6%, and 30.4% for precision, intermediate, and power grasps respectively [26]. This difference may be due to object and task selection and the additional category of intermediate grasps from [26].

### D. Object Geometry

Section II. B. divides the grasped objects into three categories: SoRs, cuboids, and irregular objects. It is hypothesized that the range of $\varphi$ should be larger for irregular objects than cuboids than SoRs. Table I displays the standard deviation for the hand position variables. The $\theta$ variable is consistent across object groups while the $\varphi$ and $r$ STDs agree less. The $r$ variable parameterizes the distance from the wrist to the origin; thus its STD implies the variation of object size in a group. The SoRs contain some of the largest and smallest objects: plate, water bottle, teacup, sprayer bottle. In contrast, object size differences in cuboid and irregular object groups are less apparent, which justifies the diverse STD for $r$. Unlike $r$, the STD for $\varphi$ grows from SoRs (12.12, 12.72) to cuboids (17.48, 15.74) to irregular objects (23.94, 20.02). The axis that the objects are rotated around, the object centerline orthogonal to the platform, is also the axis which the $\varphi$ parameter rotates around. Thus, the STD of $\varphi$ across different groups is associated with the degree of rotational symmetry among a group of objects. Since the magnitude of rotational symmetry decreases from SoRs to cuboids to irregular objects, the corresponding $\varphi$ STD shall increase.

TABLE I. WRIST POSITION PARAMETER STANDARD DEVIATION

|       |   | SoRs  | Cuboid | Irregular Objects |
|-------|---|-------|--------|-------------------|
| Right | $\varphi$ | 12.12 | 17.48  | 23.94 |
|       | $\theta$ | 13.71 | 13.19  | 13.925 |
|       | $r$ | 27.10 | 13.82  | 18.75 |
| Left  | $\varphi$ | 12.72 | 15.74  | 20.02 |
|       | $\theta$ | 12.48 | 12.11  | 13.30 |
|       | $r$ | 27.20 | 14.42  | 19.88 |

## IV. DISCUSSION AND CONCLUSION

This paper investigated a hypothesis inspired by an observation of human grasp patterns. With a collection of 13,440 grasps, we verify that human-planned grasps are subject to relatively little change when grasping objects with different orientations. We also discover that a small subset of grasp types can stably grasp most ordinary objects, and that the transformation parameters of the grasps across different graspers, objects, and orientations tend to follow the normal distribution when examined independently. Our results on the grasp type frequency qualitatively agrees with that of Cini et al. [26], but not down to specific numbers. We suspect the higher proportion of precision grasps that we received resulted from the long duration of our experiments. It takes roughly two hours for each subject to perform the 480 grasps, and it is possible that subjects become exhausted of selecting the optimal grasp for a novel object after a certain point and resort to grasps that have worked in the past, which mostly contains precision grasps.

In practice, feasibility of a grasp is a key factor to consider. Existing models either restrict the planner in the agent's workspace or let the planner run un-restricted then filter the executable grasps. A corollary from Fig. 5 and Table II. is that the latter approach, including all orientations when searching for the approach vector, is largely unnecessary. Data from the GUVOOD indicates that when the objects are rotated 360 degrees through their centerline orthogonal to the table, only around 120 degrees (smaller for SoRs and larger for Irregular Objects) of variation in the approach vector would cover 99% of successful grasps' variation. With our results, one can first determine the optimal approach vector for the robot, then find the search space using confidence intervals from the GUVOOD. This shall improve the efficiency of finding grasps that are not only stable but also executable. For non-anthropomorphic hands, scaling the intervals based on the size, dexterity of the hand, and specific tasks may be necessary. The hand-object distance statistics from this paper indicates 99% of the recorded grasps are between 70 and 210mm away from the origin. This would make the interval chosen in the study of Bohg et al. (0-200mm) a fairly accurate estimate, and the distance interval in GraspIt! (0-433mm) unnecessarily high for anthropomorphic hands. The hand-object distance results from this study may provide a reference for setting such bounds in future studies as well.

TABLE II. CONFIDENCE INTERVALS OF THE WRIST TRANSFORMATION PARAMETERS

|  |  | mean | 99% Interval | | 95% Interval | | 90% Interval | |
|---|---|---|---|---|---|---|---|---|
|  |  |  | Lower | upper | lower | upper | lower | upper |
| Right Hand | $\varphi$ | 48.06 | -22.27 | 113.07 | 8.86 | 85.17 | 16.95 | 77.55 |
|  | $\theta$ | 46.45 | 9.09 | 80.63 | 16.45 | 70.75 | 21.37 | 67.36 |
|  | $r$ | 116.30 | 75.51 | 209.60 | 84.56 | 168.68 | 89.38 | 154.43 |
|  | $Rx$ | -80.63 | -118.61 | -39.73 | -108.32 | -50.10 | 103.71 | -55.52 |
|  | $Ry$ | 1.15 | -24.17 | 33.47 | -18.52 | 26.66 | -15.85 | 22.86 |
|  | $Rz$ | -25.97 | -94.19 | 21.46 | -78.81 | 12.00 | -68.89 | 6.62 |
| Left Hand | $\varphi$ | -42.92 | -95.75 | 14.76 | -75.97 | -7.36 | -68.67 | -14.17 |
|  | $\theta$ | 50.49 | 13.17 | 77.02 | 21.32 | 73.19 | 26.16 | 70.87 |
|  | $r$ | 122.05 | 79.33 | 212.86 | 89.63 | 184.16 | 95.01 | 162.166 |
|  | $Rx$ | 89.89 | 52.02 | 126.42 | 61.49 | 115.64 | 66.23 | 110.94 |
|  | $Ry$ | 29.40 | -19.11 | 85.85 | -9.48 | 73.26 | -4.64 | 66.24 |
|  | $Rz$ | -3.48 | -35.80 | 21.01 | -26.56 | 15.14 | -21.34 | 11.89 |

In this paper, we describe our methodology to obtain the GUVOOD and preliminarily analyze the dataset. However, this does not mean that further analysis of the dataset cannot be done. When we presented the distribution of the parameters in Fig. 5, each parameter is analyzed independently. But it is unlikely that the parameters are independent of each other. Future work involving the GUVOOD could include analyzing the correlation between the parameters. Furthermore, deriving grasp planning protocols using machine learning and algebraic methods with this dataset may also be relevant future work. The dataset that we gathered in this study is available at https://github.com/dcheng728/The-GUVOOD.

ACKNOWLEDGMENT

The experimental protocol was established, according to the ethical guidelines of the Helsinki Declaration and was approved by the Human Ethics Committee of Beihang University.